\documentclass{article}

\usepackage{times}
\usepackage{graphicx} % more modern
\usepackage{subcaption}
\usepackage{pdflscape}
\usepackage{tikz}
\usetikzlibrary{calc}
\usepackage{natbib}
\usepackage{amsfonts}
\usepackage{algorithm}
\usepackage{algorithmic}
\usepackage{colortbl}
\usepackage{pgfplots}
\usepackage{amsmath}    % need for subequations
\usepackage{graphicx}   % need for figures
\usepackage{verbatim}   % useful for program listings
\usepackage{color}      % use if color is used in text
\usepackage{siunitx}
\usepackage{pgfplots}
\usepackage{pgfplotstable}

\usetikzlibrary{arrows}
\usepackage[accepted]{icml2016} 

\icmltitlerunning{A New Hierarchical Redundancy Eliminated Tree Augmented Na\"{i}ve Bayes Classifier for Coping with Gene Ontology-based Features}

\begin{document} 

\twocolumn[
\icmltitle{A New Hierarchical Redundancy Eliminated Tree Augmented Na\"{i}ve Bayes Classifier for Coping with Gene Ontology-based Features}

% It is OKAY to include author information, even for blind
% submissions: the style file will automatically remove it for you
% unless you've provided the [accepted] option to the icml2016
% package.
\icmlauthor{Cen Wan$^{\dagger,\ddagger}$}{C.Wan@ucl.ac.uk}
\icmlauthor{Alex A. Freitas$^{\ddagger}$}{A.A.Freitas@kent.ac.uk}
\icmladdress{$^\dagger$ Department of Computer Science, University College London, London, United Kingdom\\$^\ddagger$ School of Computing, Univerisity of Kent, Canterbury, United Kingdom}

% You may provide any keywords that you 
% find helpful for describing your paper; these are used to populate 
% the "keywords" metadata in the PDF but will not be shown in the document
\icmlkeywords{boring formatting information, machine learning, ICML}

\vskip 0.3in
]

\begin{abstract} 
The Tree Augmented Na\"{i}ve Bayes classifier is a type of probabilistic graphical model that can represent some feature dependencies. In this work, we propose a Hierarchical Redundancy Eliminated Tree Augmented Na\"{i}ve Bayes (HRE--TAN) algorithm, which considers removing the hierarchical redundancy during the classifier learning process, when coping with data containing hierarchically structured features. The experiments showed that HRE--TAN obtains significantly better predictive performance than the conventional Tree Augmented Na\"{i}ve Bayes classifier, and enhanced the robustness against imbalanced class distributions, in aging-related gene datasets with Gene Ontology terms used as features. 
\end{abstract}

\section{Introduction}
This work proposes a new type of Tree Augmented Na\"{i}ve Bayes (TAN) classifier, namely the Hierarchical Redundancy Eliminated Tree Augmented Na\"{i}ve Bayes (HRE--TAN) algorithm, which is designed for coping with features organized into a hierarchy (e.g., a tree or a DAG -- directed acyclic graph). In this paper the features are DAG-structured Gene Ontology (GO) terms, in datasets where instances represent genes to be classified into pro-longevity or anti-longevity genes. However, the proposed algorithm can also be applied to other classification datasets with hierarchical features.\newline\newline
Tree Augmented Na\"{i}ve Bayes (TAN) is a type of semi-Na\"{i}ve Bayes classifier that relaxes Na\"{i}ve Bayes' feature independence assumption, by allowing each feature to depend on at most one non-class variable feature. This type of tree structure-based Bayesian classifier is able to represent some feature dependencies and scales to large datasets more efficiently than other Bayesian classifiers that represent more complex feature dependencies. In this work, we focus on one of the most computationally efficient TAN classifiers \citep{Friedman1997, Keogh1999,Jiang2005TAN,Zhang2001TAN}, which essentially computes the conditional mutual information (CMI) for each pair of features given the class attribute and then builds a Maximum Weight Spanning Tree (MST), where an edge's weight is given by its CMI \citep{Friedman1997}. Then, a randomly selected vertex of the MST acts as the tree's root, and the edge directions are propagated accordingly.  

\section{Background}
\subsection{The Gene Ontology and Hierarchical Redundancy}
The Gene Ontology (GO) uses unified and structured vocabularies to describe gene functions \citep{GO2000}. Most GO terms are hierarchically structured by an ``is-a'' relationship, where each GO term is a specialization of its ancestor (more generic) terms. For example, GO:0003674 (molecular function) is the root of the DAG for molecular function terms, and it is also the parent of GO:0003824 (catalytic activity), which is in turn the parent of GO:0004803 (transposase activity). \newline\newline
This feature hierarchy has two types of hierarchical redundancy. First, if a GO term (feature) takes the value ``1'' for a given instance (gene), this implies its ancestor terms in the GO DAG also take the value ``1'' for that instance. Conversely, if the GO term takes the value ``0'' for a given instance, its descendants in the DAG also take the value ``0'' for that instance. In order to cope with those types of hierarchical redundancy, in our previous works \citep{Wan2013,Wan2014,WANPHD,Wan2016AIMatters}, three types of filter hierarchical feature selection algorithms were proposed, i.e., MR, HIP and the hybrid HIP-MR. Those three algorithms eliminate/alleviate the above types of hierarchical redundancy in a data pre-processing phase, before learning the classifier. In contrast, the proposed HRE--TAN eliminates the hierarchical redundancy during the classifier learning phase.
\subsection{Lazy Learning}
A ``lazy'' learning method performs the learning process in the testing phase, building a specific classification model for each testing instance to be classified \citep{Aha1997, Pereira2011}, rather than building a general classifier for all testing instances. The newly proposed TAN classifier in this work is based on lazy learning, since it selects features for each testing instance separately.

\section{Hierarchical Redundancy Eliminated Tree Augmented Na\"{i}ve Bayes (HRE--TAN)} 
This is a new type of tree-based Bayesian classifier based on the lazy learning approach, and it performs an embedded hierarchical feature redundancy elimination, rather than in a pre-processing step. As mentioned in Section 1, a conventional TAN method builds a MST to detect dependencies among features, but it assumes that the features are ``flat'', not hierarchical. In contrast, the proposed method eliminates the hierarchical redundancy between features when it builds the MST for each testing instance. As discussed in Section 2.1, two vertices are hierarchically redundant if one of them is an ancestor or descendant of the other and they have the same feature value (``1'' or ``0''). In essence, HRE--TAN checks the status of each edge before adding it into the Undirected Acyclic Graph (UDAG) that will be transformed into the MST later. The status of an edge will be set to ``Unavailable'' if either of the vertices connected by the edge is hierarchically redundant, with respect to the vertices that have already been included in the UDAG. The pseudocode of HRE--TAN is described in Algorithms 1 and 2.\newline\newline
In Algorithm 1, in the first part of the HRE--TAN algorithm (lines 1--12), HRE--TAN firstly generates the Directed Acyclic Graph (DAG) for the current dataset with a corresponding set of vertices (features) $\mathbb{X}$ and set of edges $\mathbb{E}$. Then it generates the set of ancestor and descendant features for each feature $x_i$, denoted $\mathbb{A}(x_i)$ and $\mathbb{D}(x_i)$, respectively. $\mathbf{Status}_{<E>}(x_i, x_j)$, which is initialized as ``Available'', denotes the selection status of the edge connecting vertices $x_i$ and $x_j$. $\mathbf{CMI}_{<E>}(x_i, x_j)$ denotes the value of CMI (conditional mutual information) for the edge $\mathbf{E}(x_i, x_j)$. All edges are sorted in descending order of their CMI value $($a greater CMI value means a higher priority of adding the edge into the UDAG$)$. In the second part of the HRE--TAN algorithm (lines 13--21), the tree $\mathbb{T}$ will be built for each testing instance (adopting a lazy learning approach) by calling the procedure HRE--MST() for building the Hierarchical Redundancy Eliminated Maximum Weight Spanning Tree (HRE--MST). Then the $\mathbf{TrainSet}$ and the current testing instance $\mathbf{Inst}_{<w>}$ will be re-created with the set $\mathbb{X}^{'}$ of the features included in the tree, so that only those features will be used for classifying the re-created testing instance. The re-created $\mathbf{TrainSet}\_\mathbf{T}$ and $\mathbf{Inst}\_\mathbf{T}_{<w>}$ with tree $\mathbb{T}$ are then used to build a lazy TAN model that classifies $\mathbf{Inst}_{<w>}$ in line 17. Finally, in lines 18--20 all edges in the DAG have their status re-assigned to ``Available'', as a preparation to process the next testing instance.\newline

\begin{algorithm}[H]
{\fontsize{9pt}{14.5pt}\selectfont
\caption{\bf{\small Lazy Hierarchical Redundancy Eliminated Tree Augmented Na\"{i}ve Bayes (HRE--TAN)}} 
\begin{algorithmic}[1]
\STATE Initialize $\mathbf{DAG}$ with all features in Dataset;
\STATE Initialize $\mathbf{TrainSet}$;
\STATE Initialize $\mathbf{TestSet}$;
\FOR {each feature $x_i$ $\in$ $\mathbb{X}$}
\STATE Initialize $\mathbb{A}(x_i)$ in $\mathbf{DAG}$;
\STATE Initialize $\mathbb{D}(x_i)$ in $\mathbf{DAG}$;
\ENDFOR
\FOR {each $\mathbf{E}(x_i, x_j)$ $\in$ $\mathbb{E}$}
\STATE Calculate $\mathbf{CMI}_{<E>}(x_i, x_j)$ using $\mathbf{TrainSet}$;
\STATE Initialize $\mathbf{Status}_{<E>}(x_i, x_j) \gets $ ``\emph{Available}'';
\ENDFOR
\STATE Sort all $\mathbf{E}(x_i, x_j)$ $\in$ $\mathbb{E}$ by descending order of $\mathbf{CMI}$;
\FOR {each instance $\mathbf{Inst}_{<w>}$ $\in$ $\mathbf{TestSet}$}
\STATE $\mathbb{T}$ = HRE--MST$($$\mathbf{DAG}$, $\mathbf{Inst}_{<w>}$, $\mathbb{A}(\mathbb{X})$, $\mathbb{D}(\mathbb{X})$, $\mathbb{E}$$)$;
\STATE Re-create $\mathbf{TrainSet}\_\mathbf{T}$ with feature set $\mathbb{X}^{'} \in$ $\mathbb{T}$;
\STATE Re-create $\mathbf{Inst}\_\mathbf{T}_{<w>}$ with feature set $\mathbb{X}^{'} \in$ $\mathbb{T}$;
\STATE Classify by TAN$(\mathbb{T}, \mathbf{TrainSet}\_\mathbf{T}, \mathbf{Inst}\_\mathbf{T}_{<w>})$;
\FOR {each $\mathbf{E}(x_i, x_j)$ $\in$ $\mathbb{E}$}
\STATE Re-assign $\mathbf{Status}_{<E>}(x_i, x_j) \gets $ ``\emph{Available}'';
\ENDFOR
\ENDFOR
\end{algorithmic}
}
\end{algorithm}

\begin{figure*}[!t]
\centering
\renewcommand{\arraystretch}{1.6}
\resizebox{12cm}{!}{
\begin{tikzpicture}[>=stealth',shorten >=1pt,auto,node distance=3cm,
                    thick,main node/.style={circle,draw,font=\sffamily\bfseries}]

  \node[main node] (1) {\Huge F};
  \node[main node] (2) [below of=1, node distance=3cm] {\Huge C};
  \node[main node] (3) [below of=2, node distance=3cm] {\Huge D};
  \node[main node] (4) [left of=2, node distance=3cm] {\Huge B};
  \node[main node] (5) [right of=2, node distance=3cm] {\Huge A};
  \node[main node] (6) [right of=1, node distance=3.0cm] {\Huge E};

 \node[main node,,draw=none] (8) [above of=6, node distance=5.6cm] {};
  \node[main node,,draw=none] (100) [right of=8, node distance=4.5cm] {};
 \node[main node] (9) [right of=100, node distance=1cm] {\Huge{F}};
\node[main node] (10) [right of=9, node distance=2.5cm] {\Huge{A}};
 \node[main node] (11) [below of=9, node distance=1.4cm] {\Huge{E}};
\node[main node] (12) [right of=11, node distance=2.5cm] {\Huge{D}};
 \node[main node] (13) [below of=11, node distance=1.4cm] {\Huge{C}};
\node[main node] (14) [right of=13, node distance=2.5cm] {\Huge{E}};
 \node[main node] (15) [below of=13, node distance=1.4cm] {\Huge{C}};
\node[main node] (16) [right of=15, node distance=2.5cm] {\Huge{D}};
 \node[main node] (17) [below of=15, node distance=1.4cm] {\Huge{B}};
\node[main node] (18) [right of=17, node distance=2.5cm] {\Huge{D}};
 \node[main node] (19) [below of=17, node distance=1.4cm] {\Huge{F}};
\node[main node] (20) [right of=19, node distance=2.5cm] {\Huge{C}};
 \node[main node] (21) [below of=19, node distance=1.4cm] {\Huge{F}};
\node[main node] (22) [right of=21, node distance=2.5cm] {\Huge{B}};
 \node[main node] (23) [below of=21, node distance=1.4cm] {\Huge{B}};
\node[main node] (24) [right of=23, node distance=2.5cm] {\Huge{E}};
 \node[main node] (25) [below of=23, node distance=1.4cm] {\Huge{B}};
\node[main node] (26) [right of=25, node distance=2.5cm] {\Huge{C}};
 \node[main node] (27) [below of=25, node distance=1.4cm] {\Huge{B}};
\node[main node] (28) [right of=27, node distance=2.5cm] {\Huge{A}};
 \node[main node] (29) [below of=27, node distance=1.4cm] {\Huge{C}};
\node[main node] (30) [right of=29, node distance=2.5cm] {\Huge{A}};
 \node[main node] (31) [below of=29, node distance=1.4cm] {\Huge{F}};
\node[main node] (32) [right of=31, node distance=2.5cm] {\Huge{E}};
 \node[main node] (33) [below of=31, node distance=1.4cm] {\Huge{A}};
\node[main node] (34) [right of=33, node distance=2.5cm] {\Huge{D}};
 \node[main node] (35) [below of=33, node distance=1.4cm] {\Huge{F}};
\node[main node] (36) [right of=35, node distance=2.5cm] {\Huge{D}};
 \node[main node] (37) [below of=35, node distance=1.4cm] {\Huge{E}};
\node[main node] (38) [right of=37, node distance=2.5cm] {\Huge{A}};
\node[main node,draw=none] (111) [left of=38, node distance=6.5cm] {};
\node[main node,draw=none] (112) [below of=111, node distance=2.8cm] {\Huge{(a)}};
\node[main node,draw=none] (113) [right of=112, node distance=15.5cm] {\Huge{(b)}};
\node[main node,draw=none] (114) [right of=113, node distance=15.5cm] {\Huge{(c)}};
 \node[main node,,draw=none] (50) [right of=38, node distance=0.65cm] {};
  \node[main node,,draw=none] (51) [below of=50, node distance=1cm] {};
  \node[main node,,draw=none] (52) [left of=51, node distance=17cm] {};
  \node[main node,,draw=none] (53) [above of=51, node distance=14.5cm] {};
  \node[main node,,draw=none] (54) [left of=53, node distance=17cm] {};
\node[main node,,draw=none] (7) [above right of=9, node distance=2.0cm] {\Huge{Sorted Edges:}};

\node[main node,draw=none] (v1) [right of=1,node distance=1.2cm] {\Huge \textbf{1}};
\node[main node,draw=none] (v2) [right of=2,node distance=1.2cm] {\Huge \textbf{1}};
\node[main node,draw=none] (v3) [right of=3,node distance=1.2cm] {\Huge \textbf{0}};
\node[main node,draw=none] (v4) [right of=4,node distance=1.2cm] {\Huge \textbf{0}};
\node[main node,draw=none] (v5) [right of=5,node distance=1.2cm] {\Huge \textbf{0}};
\node[main node,draw=none] (v6) [right of=6,node distance=1.2cm] {\Huge \textbf{1}};

\draw[-triangle 45](1) -- (2);
\draw[-triangle 45](1) -- (4);
\draw[-triangle 45](2) -- (3);
\draw[-triangle 45](5) -- (3);
\draw[-triangle 45](6) -- (5);

\path
(9) edge (10)
 (10)  edge  (9) 
 (11)  edge  (12) 
 (13)  edge  (14) 
 (14)  edge  (13) 
 (15)  edge  (16) 
 (17)  edge  (18) 
 (18)  edge  (17) 
 (19)  edge  (20) 
 (21)  edge  (22) 
 (23)  edge  (24) 
 (24)  edge  (23) 
 (25)  edge  (26) 
 (26)  edge  (25) 
 (27)  edge  (28) 
 (28)  edge  (27) 
 (29)  edge  (30) 
 (30)  edge  (29) 
 (31)  edge  (32) 
 (32)  edge  (31) 
 (33)  edge  (34) 
 (35)  edge  (36) 
 (37)  edge  (38);

\node[main node,draw=none] (Arrow) [right of=22,node distance=2.95cm] {\Huge$\Longrightarrow$};

  \node[main node] (B1) [fill=green,right of=1, node distance=20cm] {\Huge F};
  \node[main node] (B2) [fill=red, right of=2, node distance=20cm] {\Huge \color{white}C};
  \node[main node] (B3) [fill=red, right of=3, node distance=20cm] {\Huge \color{white}D};
  \node[main node] (B4) [fill=green,right of=4, node distance=20cm] {\Huge B};
  \node[main node] (B5) [fill=green,right of=5, node distance=20cm] {\Huge A};
  \node[main node] (B6) [fill=green,right of=6, node distance=20cm] {\Huge E};

 \node[rectangle,fill=green,below of=B3,node distance=8cm] {\Huge\bf \hspace{0.18cm}Selected\color{green}.};
 \node[rectangle,fill=red,below of=B3,node distance=6.0cm] {\Huge\hspace{0.04cm}\bf\color{white}Removed};
 
 \node[main node,,draw=none] (B8) [right of=8, node distance=20cm] {};
  \node[main node,,draw=none] (B100) [right of=100, node distance=20cm] {};
 \node[main node] (B9) [fill=green, right of=9, node distance=20cm] {\Huge{F}};
\node[main node] (B10) [fill=green, right of=10, node distance=20cm] {\Huge{A}};
 \node[main node] (B11) [fill=red,right of=11, node distance=20cm] {\color{white}\Huge{E}};
\node[main node] (B12) [fill=red,right of=12, node distance=20cm] {\color{white}\Huge{D}};
 \node[main node] (B13) [fill=red,right of=13, node distance=20cm] {\color{white}\Huge{C}};
\node[main node] (B14) [fill=red,right of=14, node distance=20cm] {\color{white}\Huge{E}};
 \node[main node] (B15) [fill=red,right of=15, node distance=20cm] {\color{white}\Huge{C}};
\node[main node] (B16) [fill=red,right of=16, node distance=20cm] {\color{white}\Huge{D}};
 \node[main node] (B17) [fill=red,right of=17, node distance=20cm] {\color{white}\Huge{B}};
\node[main node] (B18) [fill=red,right of=18, node distance=20cm] {\color{white}\Huge{D}};
 \node[main node] (B19) [fill=red,right of=19, node distance=20cm] {\color{white}\Huge{F}};
\node[main node] (B20) [fill=red,right of=20, node distance=20cm] {\color{white}\Huge{C}};
 \node[main node] (B21) [fill=green,right of=21, node distance=20cm] {\Huge{F}};
\node[main node] (B22) [fill=green,right of=22, node distance=20cm] {\Huge{B}};
 \node[main node] (B23) [fill=green,right of=23, node distance=20cm] {\Huge{B}};
\node[main node] (B24) [fill=green,right of=24, node distance=20cm] {\Huge{E}};
 \node[main node] (B25) [fill=red,right of=25, node distance=20cm] {\color{white}\Huge{B}};
\node[main node] (B26) [fill=red,right of=26, node distance=20cm] {\color{white}\Huge{C}};
 \node[main node] (B27) [fill=red,right of=27, node distance=20cm] {\color{white}\Huge{B}};
\node[main node] (B28) [fill=red,right of=28, node distance=20cm] {\color{white}\Huge{A}};
 \node[main node] (B29) [fill=red,right of=29, node distance=20cm] {\color{white}\Huge{C}};
\node[main node] (B30) [fill=red,right of=30, node distance=20cm] {\color{white}\Huge{A}};
 \node[main node] (B31) [fill=red,right of=31, node distance=20cm] {\color{white}\Huge{F}};
\node[main node] (B32) [fill=red,right of=32, node distance=20cm] {\color{white}\Huge{E}};
 \node[main node] (B33) [fill=red,right of=33, node distance=20cm] {\color{white}\Huge{A}};
\node[main node] (B34) [fill=red,right of=34, node distance=20cm] {\color{white}\Huge{D}};
 \node[main node] (B35) [fill=red,right of=35, node distance=20cm] {\color{white}\Huge{F}};
\node[main node] (B36) [fill=red,right of=36, node distance=20cm] {\color{white}\Huge{D}};
 \node[main node] (B37) [fill=red,right of=37, node distance=20cm] {\color{white}\Huge{E}};
\node[main node] (B38) [fill=red,right of=38, node distance=20cm] {\color{white}\Huge{A}};
 \node[main node,,draw=none] (B50) [right of=50, node distance=20cm] {};
  \node[main node,,draw=none] (B51) [right of=51, node distance=20cm] {};
  \node[main node,,draw=none] (B52) [right of=52, node distance=20cm] {};
  \node[main node,,draw=none] (B53) [right of=53, node distance=20cm] {};
  \node[main node,,draw=none] (B54) [right of=54, node distance=20cm] {};
\node[main node,,draw=none] (B7) [right of=7, node distance=20cm] {\Huge{Sorted Edges:}};

\node[main node,draw=none] (Bv11) [right of=v1,node distance=20cm] {\Huge \textbf{1}};
\node[main node,draw=none] (Bv12) [right of=v2,node distance=20cm] {\Huge \textbf{1}};
\node[main node,draw=none] (Bv13) [right of=v3,node distance=20cm] {\Huge \textbf{0}};
\node[main node,draw=none] (Bv14) [right of=v4,node distance=20cm] {\Huge \textbf{0}};
\node[main node,draw=none] (Bv15) [right of=v5,node distance=20cm] {\Huge \textbf{0}};
\node[main node,draw=none] (Bv16) [right of=v6,node distance=20cm] {\Huge \textbf{1}};

\draw[-triangle 45](B1) -- (B2);
\draw[-triangle 45](B1) -- (B4);
\draw[-triangle 45](B2) -- (B3);
\draw[-triangle 45](B5) -- (B3);
\draw[-triangle 45](B6) -- (B5);

\path
(B9) edge (B10)
 (B10)  edge  (B9) 
 (B11)  edge  (B12) 
 (B13)  edge  (B14) 
 (B14)  edge  (B13) 
 (B15)  edge  (B16) 
 (B17)  edge  (B18) 
 (B18)  edge  (B17) 
 (B19)  edge  (B20) 
 (B21)  edge  (B22) 
 (B23)  edge  (B24) 
 (B24)  edge  (B23) 
 (B25)  edge  (B26) 
 (B26)  edge  (B25) 
 (B27)  edge  (B28) 
 (B28)  edge  (B27) 
 (B29)  edge  (B30) 
 (B30)  edge  (B29) 
 (B31)  edge  (B32) 
 (B32)  edge  (B31) 
 (B33)  edge  (B34) 
 (B35)  edge  (B36) 
 (B37)  edge  (B38);

\node[main node,draw=none] (Arrow) [right of=B22,node distance=2.95cm] {\Huge$\Longrightarrow$};
  
  \node[main node,,draw=none] (C0) [right of=B22, node distance=3.6cm] {};
  \node[main node,,draw=none] (C100) [below of=C0, node distance=0.6cm] {};
  \node[main node] (C1) [above right of=C100, node distance=2.8cm] {\Huge F};
  \node[main node] (C2) [right of=C1, node distance=2.8cm] {\Huge B};
  \node[main node] (C3) [below of=C1, node distance=2.8cm] {\Huge A};
  \node[main node] (C4) [below of=C2, node distance=2.8cm] {\Huge E};
  \node[main node,draw=none] (C5) [below right of=C2, node distance=2cm] {};
  \node[main node] (C6) [right of=C5, node distance=1cm] {\Huge{Class}};
  \node[main node,draw=none] (C11) [below of=C6, node distance=3cm] {};

  \draw[-triangle 45](C2) -- (C1);
  \draw[-triangle 45](C2) -- (C4);
  \draw[-triangle 45](C1) -- (C3);
  \draw[dashed,-triangle 45](C6) -- (C1);
  \draw[dashed,-triangle 45](C6) -- (C2);
  \draw[dashed,-triangle 45](C6) -- (C3);
  \draw[dashed,-triangle 45](C6) -- (C4);

\end{tikzpicture}
}
\caption{Example of HRE--TAN operation initially with a set of features structured as a DAG}
\end{figure*}
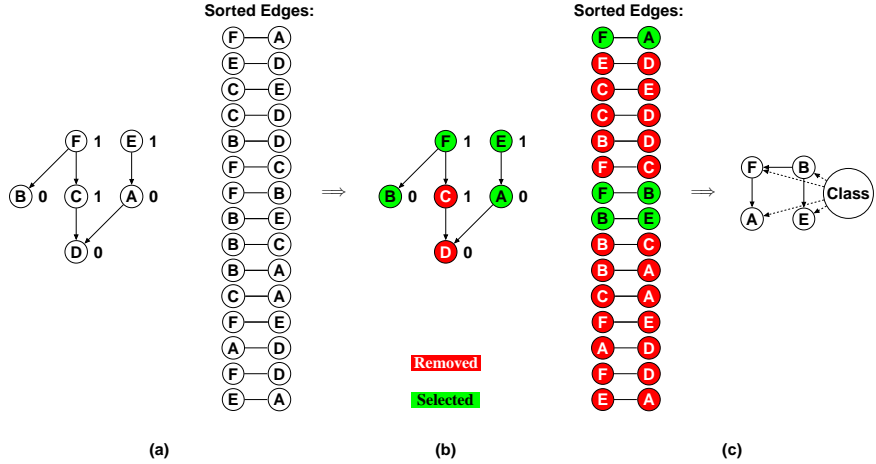

Algorithm 2 shows the pseudocode for building the HRE--MST. $\mathbf{NR}$($x_i$, $x_j$, $\mathbf{Inst}_{<w>}$, $\mathbf{DAG}$) is a Boolean function that returns ``True'' if nodes $x_i$ and $x_j$ are non-hierarchically-redundant in the current testing instance $\mathbf{Inst}_{<w>}$, given the feature DAG. $\mathbf{NoCycle}(\mathbf{E}(x_i, x_j), \mathbf{UDAG})$ is a Boolean function that returns ``True'' if there is no cycle in the $\mathbf{UDAG}$ after adding edge $\mathbf{E}(x_i, x_j)$. If the edge satisfies all the conditions in line 3 of Algorithm 2, it will be added into the $\mathbf{UDAG}$ (line 4). Once the algorithm has added the edge $\mathbf{E}(x_i, x_j)$ to the $\mathbf{UDAG}$, for each of the two nodes connected by that edge, denoted as $x_g$ (line 5), the algorithm will consider each of the nodes which is either an ancestor or a descendant of $x_g$ in the feature $\mathbf{DAG}$, denoting each such ancestor/descendant as $x_h$ (line 6). If feature $x_g$ and its ancestor/descendant feature $x_h$ have the same value in the current testing instance $\mathbf{Inst}_{<w>}$ (line 7), indicating a hierarchical redundancy in that pair of features, then the \emph{for each} loop in lines 8--10 will set to ``Unavailable'' the status of all edges where one of the nodes is $x_h$ -- line 8, where the symbol ``$\ast$'' is a wildcard matching any node. In other words, among the set of hierarchically-redundant nodes (features) with the same value, HRE--TAN selects the node included in the edge having higher conditional mutual information (CMI), since Algorithm 2 processes edges in descending order of CMI. \newline

To explain how Algorithms 1 and 2 work, we use the example DAG shown in Figure 1.a, where the left part is a feature hierarchy consisting of three paths from a root to a leaf node of the $\mathbf{DAG}$, i.e., node F to node B; node F to node D; and node E to node D. The right part of Figure 1.a shows the edges (for all pair of nodes) in descending order of $\mathbf{CMI}$. HRE--TAN firstly adds edge $\mathbf{E}(F, A)$ into the UDAG, since its selection status is ``\emph{Available}''; nodes F and A are not hierarchically-redundant; and there is no cycle in the $\mathbf{UDAG}$ after adding edge $\mathbf{E}(F, A)$. Then, Algorithm 2 will delete all edges that contain hierarchically redundant nodes with respect to node F or node A, in order to minimize feature redundancy. Node C is redundant with respect to node F, because both of them have value ``1'' and are located in the same path in Figure 1.a. So, all edges containing node C $($i.e., $\mathbf{E}(C, E)$, $\mathbf{E}(C, D)$, $\mathbf{E}(F, C)$, $\mathbf{E}(B, C)$, $\mathbf{E}(C, A)$$)$ will be unavailable to be added into the $\mathbf{UDAG}$. Also, node D is redundant with respect to node A, because both of them have value ``0'' and are located in the same path. Then, all edges containing node D (i.e., $\mathbf{E}(E, D)$, $\mathbf{E}(C, D)$, $\mathbf{E}(B, D)$, $\mathbf{E}(A, D)$, $\mathbf{E}(F, D)$) will be unavailable to be added into the $\mathbf{UDAG}$. Note that this hierarchical redundancy elimination process will dramatically reduce the size of the search space of candidate TAN structures. \newline

\begin{algorithm}[H]
{\fontsize{9pt}{14.5pt}\selectfont
\caption{\bf{\small Hierarchical Redundancy Eliminated Maximum Weight Spanning Tree (HRE--MST)}} 
\begin{algorithmic}[1]
\STATE Initialize an Empty $\mathbf{UDAG}$;
 \FOR {each $\mathbf{E}(x_i, x_j)$ $\in$ $\mathbb{E}$}
  \IF {\{$\mathbf{Status}_{<E>}(x_i, x_j)$ = ``\emph{Available}''\} $\wedge$ \newline\indent\hspace{1.9mm} \{$\mathbf{NR}($$x_i$, $x_j$, $\mathbf{Inst}_{<w>}$, $\mathbf{DAG}$$)$\} $\wedge$ \newline\indent\hspace{2.7mm}\{$\mathbf{NoCycle}($$\mathbf{E}$($x_i$, $x_j$), $\mathbf{UDAG}$$)$\}}
         \STATE add $\mathbf{E}(x_i, x_j)$ into $\mathbf{UDAG}$;
         \FOR {each $x_g$ in \{$x_i$, $x_j$\}} 
             \FOR {each $x_h$ in \{$\mathbb{A}(x_g)$ $\cup$ $\mathbb{D}(x_g)$\}}
                    \IF {$\mathbf{V}(x_g, \mathbf{Inst}_{<w>}$) = $\mathbf{V}(x_h, \mathbf{Inst}_{<w>}$)}
                       \FOR {each $\mathbf{E}(x_h, \ast)$}
                          \STATE $\mathbf{Status}_{<E>}(x_h, \ast)$ $\gets$ ``\emph{Unavailable}'';
                       \ENDFOR
                    \ENDIF
             \ENDFOR
         \ENDFOR
\ENDIF
\ENDFOR
\STATE Choose $\mathbf{Root}$ by Randomly selecting vertex \emph{x} in $\mathbf{UDAG}$;
\STATE Build the tree ($\mathbb{T}$) by marking direction of all edges from the $\mathbf{Root}$ outwards to other vertices;
\STATE Return $\mathbb{T}$;

\end{algorithmic}
}
\end{algorithm}

\begin{table*}

\renewcommand{\arraystretch}{1.87}
\resizebox{17cm}{!}{
\begin{tabular}{cccccccccccccccc} 
\multicolumn{16}{c}{\Large \emph{Table 1.} Sensitivity ($\pm$ standard error), specificity ($\pm$ standard error) and GMean values obtained by HRE--TAN and TAN over 28 datasets}\\\hline
\bf                  & \multicolumn{7}{ c }{\Large\bf\emph{Caenorhabditis elegans Datasets}} && \multicolumn{7}{ c }{\Large\bf\emph{Drosophila melanogaster Datasets}} \\\cline{2-8}\cline{10-16}
\bf                  & \multicolumn{3}{ c }{\bf\large HRE-TAN} && \multicolumn{3}{ c }{\bf\large TAN} & & \multicolumn{3}{ c }{\bf\large HRE-TAN} && \multicolumn{3}{ c }{\bf\large TAN}\\\cline{2-4}\cline{6-8}\cline{10-12}\cline{14-16}
\bf       & \large\emph{Sens.} &\large \emph{Spec.} &\large \emph{GMean} & &\large \emph{Sens.} &\large \emph{Spec.} &\large \emph{GMean} &&\large \emph{Sens.} &\large \emph{Spec.} &\large \emph{GMean}&& \large\emph{Sens.} &\large \emph{Spec.} &\large \emph{GMean}\\\hline
\large\emph{BP}	&	\large	41.1 $\pm$ 2.4	&	\large	76.8 $\pm$ 2.1	&	\large	\bf56.2&	&	 \large	34.0 $\pm$ 3.2 	&	\large	79.6 $\pm$ 2.3	&	\large	52.0 	&&	\large	86.8 $\pm$ 3.2	&	\large	30.6 $\pm$ 10.2	&	\large	\bf51.5	&&	\large	92.3 $\pm$ 2.9	&	\large	19.4 $\pm$ 8.4	&	\large	42.3	\\
\large\emph{MF}	&	\large	23.1 $\pm$ 4.8	&	\large	75.3 $\pm$ 5.4	&	\large	41.7	&&	\large	37.2 $\pm$ 5.8 	&	\large	61.4 $\pm$ 5.0	&	\large	\bf47.8	&&	\large	86.8 $\pm$ 3.4	&	\large	41.2 $\pm$ 8.8	&	\large	\bf59.8	&&	\large	91.2 $\pm$ 3.3	&	\large	20.6 $\pm$ 5.0	&	\large	43.3	\\
\large\emph{CC}	&	\large	24.5 $\pm$ 3.6	&	\large	80.8 $\pm$ 3.0	&	\large	44.5	&&	\large	39.8 $\pm$ 3.0	&	\large	78.2 $\pm$ 2.2	&	\large	\bf55.8	&&	\large	75.8 $\pm$ 5.8	&	\large	28.6 $\pm$ 9.7	&	\large	46.6	&&	\large	90.3 $\pm$ 3.6	&	\large	32.1 $\pm$ 11.6	&	\large	\bf53.8	\\
\large\emph{BP+MF}	&	\large	42.3 $\pm$ 2.3	&	\large	80.0 $\pm$ 2.6	&	\large	\bf58.2	&&	\large	35.2 $\pm$ 1.9	&	\large	80.3 $\pm$ 2.2	&	\large	53.2	&&	\large	87.0 $\pm$ 3.3	&	\large	31.6 $\pm$ 6.5	&	\large	\bf52.4	&&	\large	92.4 $\pm$ 3.3	&	\large	23.7 $\pm$ 6.9	&	\large	46.8	\\
\large\emph{BP+CC}	&	\large	44.6 $\pm$ 3.0	&	\large	74.4 $\pm$ 3.6	&	\large	57.6	&&	\large	42.7 $\pm$ 3.1	&	\large	81.7 $\pm$ 2.7	&	\large	\bf59.1	&&	\large	84.6 $\pm$ 2.4	&	\large	32.4 $\pm$ 10.6	&	\large	\bf52.4	&&	\large	86.8 $\pm$ 4.0	&	\large	18.9 $\pm$ 7.6	&	\large	40.5	\\
\large\emph{MF+CC}	&	\large	32.4 $\pm$ 3.3	&	\large	79.8 $\pm$ 3.2	&	\large	50.8	&&	\large	40.6 $\pm$ 3.4	&	\large	74.4 $\pm$ 3.6	&	\large	\bf55.0	&&	\large	87.1 $\pm$ 4.4	&	\large	39.5 $\pm$ 5.5	&	\large	\bf58.7	&&	\large	90.6 $\pm$ 3.3	&	\large	31.6 $\pm$ 5.0	&	\large	53.5	\\
\large\emph{BP+MF+CC}	&	\large	44.2 $\pm$ 3.9	&	\large	79.3 $\pm$ 2.9	&	\large	\bf59.2	&&	\large	39.5 $\pm$ 2.8	&	\large	80.1 $\pm$ 2.6	&	\large	56.2	&&	\large	82.6 $\pm$ 3.4	&	\large	47.4 $\pm$ 8.7	&	\large	\bf62.6	&&	\large	92.4 $\pm$ 2.4	&	\large	18.4 $\pm$ 5.3	&	\large	41.2	\\\hline
\bf                  & \multicolumn{7}{ c }{\Large\bf\emph{Mus musculus Datasets}} && \multicolumn{7}{ c }{\Large\bf\emph{Saccharomyces cerevisiae Datasets}} \\\cline{2-8}\cline{10-16}
\bf                  & \multicolumn{3}{ c }{\bf\large HRE-TAN} && \multicolumn{3}{ c }{\bf\large TAN} & & \multicolumn{3}{ c }{\bf\large HRE-TAN} && \multicolumn{3}{ c }{\bf\large TAN}\\\cline{2-4}\cline{6-8}\cline{10-12}\cline{14-16}
\bf       & \large\emph{Sens.} &\large \emph{Spec.} &\large \emph{GMean} & &\large \emph{Sens.} &\large \emph{Spec.} &\large \emph{GMean} &&\large \emph{Sens.} &\large \emph{Spec.} &\large \emph{GMean}&& \large\emph{Sens.} &\large \emph{Spec.} &\large \emph{GMean}\\\hline
\large\emph{BP}	&	\large	86.8 $\pm$ 5.5& 	\large 	47.1 $\pm$ 4.7& 	\large	\bf63.9 	&&	\large89.7 $\pm$ 3.7 	&	\large	41.2 $\pm$ 4.9 &	\large	60.8 	&&	\large	20.0 $\pm$ 7.4 	&	\large	93.5 $\pm$ 1.7 	&	\large	\bf43.2	&&	\large		3.3 $\pm$ 3.3 	&	\large	98.9 $\pm$ 1.1&		\large 18.1 \\
\large\emph{MF}	&	\large	83.1 $\pm$ 3.3& 	\large	42.4 $\pm$ 9.3& 	\large	\bf59.4	&&	\large89.2 $\pm$ 4.0 	&	\large	33.3 $\pm$ 9.4& 	\large	54.5 	&&	\large	0.0 $\pm$ 0.0	&	\large	96.9 $\pm$ 1.7	&	\large	0.0	&&		\large	0.0 $\pm$ 0.0 	&	\large	97.7 $\pm$ 1.2& 	\large	0.0 \\
\large\emph{CC}	&	\large	86.4 $\pm$ 4.0& 	\large	41.2 $\pm$ 9.7& 	\large	\bf59.7	&&	\large75.8 $\pm$ 4.4 	&	\large	41.2 $\pm$ 8.3& 	\large	55.9 	&&	\large	12.5 $\pm$ 6.1	&		\large93.5 $\pm$ 2.9	&	\large	34.2	&&		\large	16.7 $\pm$ 7.0 	&	\large	95.9 $\pm$ 2.1& 	\large	\bf40.0 \\
\large\emph{BP+MF}	&	\large	83.8 $\pm$ 4.5& 	\large	41.2 $\pm$ 6.8 &	\large	\bf58.8	&&	\large86.8 $\pm$ 3.4 	&	\large	35.3 $\pm$ 5.4& 	\large	55.4 	&&	\large	26.7 $\pm$ 10.9	&	\large	95.8 $\pm$ 1.5	&		\large\bf50.6	&&		\large	3.3 $\pm$ 3.3 	&	\large	99.0 $\pm$ 0.7& 	\large	18.1 \\
\large\emph{BP+CC}	&		\large79.4 $\pm$ 4.9& 	\large	47.1 $\pm$ 9.7& 	\large	61.2 	&&	\large88.2 $\pm$ 3.6 	&	\large	47.1 $\pm$ 9.7& 	\large	\bf64.5 	&&	\large	26.7 $\pm$ 6.7	&	\large	94.1 $\pm$ 2.1	&	\large	\bf50.1	&&	\large		10.0 $\pm$ 5.1 	&	\large	99.0 $\pm$ 0.7& 	\large	31.5 \\
\large\emph{MF+CC}	&		\large89.7 $\pm$ 3.0 &	\large	35.3 $\pm$ 9.6& 	\large	56.3 	&&	\large88.2 $\pm$ 4.2 	&	\large	41.2 $\pm$ 10.0& 	\large	\bf60.3 	&&	\large	10.3 $\pm$ 6.1	&	\large	95.4 $\pm$ 1.9	&		\large\bf31.3	&&	\large		5.0 $\pm$ 5.0 	&	\large	98.5 $\pm$ 0.8& 	\large	22.2 \\
\large\emph{BP+MF+CC}	&		\large85.3 $\pm$ 3.7 &	\large	44.1 $\pm$ 8.9	&	\large 61.3	&&	\large91.2 $\pm$ 3.2	&	 	 \large41.2 $\pm$ 8.6 &	\large	61.3 	&&	\large	23.3 $\pm$ 7.1	&	\large	96.2 $\pm$ 1.4	&		\large\bf47.3	&&		\large	0.0 $\pm$ 0.0 	&	\large	99.0 $\pm$ 0.6&		\large 0.0 \\\hline
\multicolumn{16}{c}{}\\
\end{tabular}
}

\end{table*}

After edges with node C or D had their selection status set to ``\emph{Unavailable}'', edge $\mathbf{E}(F, B)$ -- the next one available in the sorted list -- will be added into the $\mathbf{UDAG}$, since nodes F and B are not redundant (although both of them are in the same path in Figure 1.a, their values are different), and there is no cycle in the $\mathbf{UDAG}$ after adding that edge. Node B is not redundant with respect to any other node, so no edge has its status set to ``Unavailable'' in this step. Then, $\mathbf{E}(B, E)$ will be added into the $\mathbf{UDAG}$ as the next available edge in the sorted edge list, since this edge also satisfies all conditions in line 3 of Algorithm 2. Then, $\mathbf{E}(B, A)$, $\mathbf{E}(F, E)$ and $\mathbf{E}(E, A)$ will be processed in turn. However, none of them will be added into the $\mathbf{UDAG}$, since this would create a cycle in that $\mathbf{UDAG}$. Figure 1.b shows the selection status of features after processing all edges, while green color denotes the features were kept and included in the UDAG, whereas red color denotes features were removed and not included in the UDAG. Finally, HRE--TAN randomly selects a node as the root, which is used to mark directions of all edges in order to build the MST. Figure 1.c shows the final tree classifier including all selected features, with choosing feature B as the root. After finding the HRE--MST (i.e., tree $\mathbb{T}$), the training dataset and current testing instance will be re-created, and the testing instance will be classified using the built tree (line 17 in Algorithm 1). Then the selection status of all edges will be re-assigned as ``Available'' in line 19 of Algorithm 1, as a preparation for processing the next testing instance. 

\section{Computational Experiments}
\subsection{Aging-related Genes Datasets}
We adopted the aging-related genes datasets used by our previous work \cite{WanACMBCB2015}. The datasets consist of aging-related genes as instances and 7 different types of combination of Gene Ontology terms as features, e.g., biological process (BP) terms with molecular function (MF) terms, or molecular function (MF) terms with cellular component (CC) terms. The aging-related genes information was about 4 different modal organisms, i.e., \emph{Caenorhabditis elegans} (CE), \emph{Drosophila melanogaster} (DM), \emph{Mus musculus} (MM) and \emph{Saccharomyces cerevisiae} (SC), obtained from Human Ageing Genomic Resources (HAGR) GenAge database \cite{Tacutu2013}. Therefore, in total we have 28 different datasets (4 model organisms times 7 types of GO terms combinations).

\subsection{Experimental Results and Discussion}
We conducted a head-to-head comparison between the newly proposed HRE--TAN and the conventional TAN classifier based on their predictive accuracy. We used a well-known 10-fold cross validation procedure to evaluate the predictive accuracy measured by the geometric mean (Gmean) of Sensitivity and Specificity, i.e., the square root of the product of Sensitivity and Specificity. Sensitivity denotes the proportion of positive (pro-longevity) genes correctly classified as positive; whilst Specificity denotes the ratio of negative (anti-longevity) genes correctly classified as negative. Table 1 displays the experimental results of HRE--TAN and TAN on the 28 datasets. The figures in bold denote higher GMean value among the two algorithms in each dataset. \hspace{1.5mm} Overall, HRE--TAN and TAN obtained the higher GMean value in 18 and 8 datasets, respectively, with the GMean result being a tie in the other two datasets. According to the two-tailed Wilcoxon signed-rank test, HRE--TAN significantly outperformed TAN at the 0.05 significance level.\newline\newline
Notably, the class distribution on the datasets is imbalanced. Hence, we evaluated the robustness of HRE--TAN and TAN against imbalanced class distributions, by calculating the correlation coefficient $r$ between GMean and the degree of class imbalance $\mathbf{D}$, given by: $\mathbf{D}$ (i.e., $\mathbf{D}=1-\frac{\mathbf{\#} (Minor)}{\mathbf{\#} (Major)}$, where $\mathbf{\#} (Minor)$ denotes the number of instances belonging to the minority class and $\mathbf{\#} (Major)$ denotes the number of instances belonging to the majority class. The values of $\mathbf{D}$ range from 0.234 (CE--MF dataset) to 0.856 (SC--BP+MF+CC dataset). The linear relationship between GMean and $\mathbf{D}$ values is shown in the scatter plots in Figure 2, where the red straight lines denote the fitted linear regression models. Obviously, HRE--TAN has better robustness against class imbalance than TAN, since HRE--TANÕs GMean decreases more slowly with an increase in $\mathbf{D}$ than TAN.\newline

\begin{figure}[!h]
  \begin{subfigure}[b]{0.23\textwidth}
    \centering
    \resizebox{\linewidth}{!}{
     \begin{tikzpicture}
	\begin{axis}[ymin=0, ymax=100,xlabel=Degree of Class Imbalance,ylabel=GMean-TAN]
	\addplot[mark=star,only marks] coordinates {
		(0.345,52.0)
		(0.234,47.8)
		(0.372,55.8)
		(0.374,53.2)
		(0.381,59.1)
		(0.351,55.0)
		(0.398,56.2)
		(0.604,42.3)
		(0.500,43.3)
		(0.548,53.8)
		(0.587,46.8)
		(0.593,40.5)
		(0.553,53.5)
		(0.587,41.2)
		(0.500,60.8)
		(0.492,54.5)
		(0.485,55.9)
		(0.500,55.4)
		(0.500,64.5)
		(0.500,60.3)
		(0.500,61.3)
		(0.838,18.1)
		(0.802,0.0)
		(0.805,40.0)
		(0.844,18.1)
		(0.853,31.5)
		(0.853,22.2)
		(0.856,0.0)
	};
	\addplot table[mark=,row sep=\\, y={create col/linear regression={y=Y}}] % compute a linear regression from the
    %input table
    {
        X Y\\
        		0.345 52.0\\
		0.234 47.8\\
		0.372 55.8\\
		0.374 53.2\\
		0.381 59.1\\
		0.351 55.0\\
		0.398 56.2\\
		0.604 42.3\\
		0.500 43.3\\
		0.548 53.8\\
		0.587 46.8\\
		0.593 40.5\\
		0.553 53.5\\
		0.587 41.2\\
		0.500 60.8\\
		0.492 54.5\\
		0.485 55.9\\
		0.500 55.4\\
		0.500 64.5\\
		0.500 60.3\\
		0.500 61.3\\
		0.838 18.1\\
		0.802 0.0\\
		0.805 40.0\\
		0.844 18.1\\
		0.853 31.5\\
		0.853 22.2\\
		0.856 0.0\\
    };
	\end{axis}
   \end{tikzpicture}}
 \subcaption{$r$(TAN) = -0.801}
 \end{subfigure}
 \begin{subfigure}[b]{0.23\textwidth}
    \centering
    \resizebox{\linewidth}{!}{
     \begin{tikzpicture}
	\begin{axis}[ymin=0, ymax=100,xlabel=Degree of Class Imbalance,ylabel=GMean-HRE--TAN]
	\addplot[mark=star,only marks] coordinates {
		(0.345,56.2)
		(0.234,41.7)
		(0.372,44.5)
		(0.374,58.2)
		(0.381,57.6)
		(0.351,50.8)
		(0.398,59.2)
		(0.604,51.5)
		(0.500,59.8)
		(0.548,46.6)
		(0.587,52.4)
		(0.593,52.4)
		(0.553,58.7)
		(0.587,62.6)
		(0.500,63.9)
		(0.492,59.4)
		(0.485,59.7)
		(0.500,58.8)
		(0.500,61.2)
		(0.500,56.3)
		(0.500,61.3)
		(0.838,43.2)
		(0.802,0.0)
		(0.805,34.2)
		(0.844,50.6)
		(0.853,50.1)
		(0.853,31.3)
		(0.856,47.3)
	};
	\addplot table[mark=,row sep=\\, y={create col/linear regression={y=Y}}]     {
        X Y\\
        0.345 56.2\\
		0.234 41.7\\
		0.372 44.5\\
		0.374 58.2\\
		0.381 57.6\\
		0.351 50.8\\
		0.398 59.2\\
		0.604 51.5\\
		0.500 59.8\\
		0.548 46.6\\
		0.587 52.4\\
		0.593 52.4\\
		0.553 58.7\\
		0.587 62.6\\
		0.500 63.9\\
		0.492 59.4\\
		0.485 59.7\\
		0.500 58.8\\
		0.500 61.2\\
		0.500 56.3\\
		0.500 61.3\\
		0.838 43.2\\
		0.802 0.0\\
		0.805 34.2\\
		0.844 50.6\\
		0.853 50.1\\
		0.853 31.3\\
		0.856 47.3\\
     };   
	\end{axis}
   \end{tikzpicture}}
 \subcaption{$r$(HRE--TAN) = -0.479}
 \end{subfigure}
 \caption{Linear relationship between $\mathbf{D}$ and GMean Values}
\end{figure}
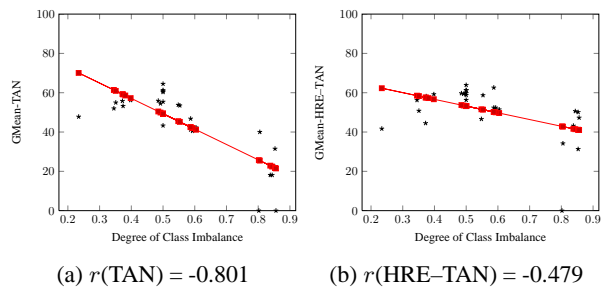

\section{Conclusion and Future Research Directions} 
In this work, we proposed and evaluated a new type of TAN classifier, i.e., HRE--TAN, which considers eliminating the hierarchical redundancy between features (hierarchical Gene Ontology terms) during the construction of the tree of features that is used as part of the classifier. The experiments show that HRE--TAN significantly outperforms the conventional TAN classifier on the tasks of classifying aging-related genes into pro-longevity or anti-longevity genes. In future work, we will further evaluate the performance of HRE--TAN in other datasets of hierarchical features and exploit other criteria to eliminate hierarchical redundancy during the classifier learning phase.
\section*{Acknowledgements} 
We thank Dr. Jo\~{a}o~Pedro~de~Magalh\~{a}es for his valuable general advice for this project.
\nocite{langley00}

\bibliography{mybibfile2}

\begin{thebibliography}{13}
\providecommand{\natexlab}[1]{#1}
\providecommand{\url}[1]{\texttt{#1}}
\expandafter\ifx\csname urlstyle\endcsname\relax
  \providecommand{\doi}[1]{doi: #1}\else
  \providecommand{\doi}{doi: \begingroup \urlstyle{rm}\Url}\fi

\bibitem[{Aha}(1997)]{Aha1997}
{Aha}, D.~W.
\newblock \emph{{Lazy Learning}}.
\newblock Kluwer Academic Publishers, Norwell, MA, 1997.

\bibitem[Friedman et~al.(1997)Friedman, Geiger, and Goldszmidt]{Friedman1997}
Friedman, N., Geiger, D., and Goldszmidt, M.
\newblock Bayesian network classifiers.
\newblock \emph{Machine Learning}, 29\penalty0 (2-3):\penalty0 131--163,
  November 1997.

\bibitem[Jiang et~al.(2005)Jiang, Zhang, Cai, and Su]{Jiang2005TAN}
Jiang, L., Zhang, H., Cai, Z., and Su, J.
\newblock Learning tree augmented naive bayes for ranking.
\newblock \emph{Database Systems for Advanced Applications}, pp.\  688--698,
  January 2005.

\bibitem[Keogh \& Pazzani(1999)Keogh and Pazzani]{Keogh1999}
Keogh, E.~J. and Pazzani, M.~J.
\newblock Learning augmented bayesian classifiers: A comparison of
  distribution-based and classification-based approaches.
\newblock In \emph{Proc. the seventh international workshop on artificial
  intelligence and statistics}, pp.\  225--230, Florida, USA, January 1999.

\bibitem[Pereira et~al.(2011)Pereira, Plastino, Zadrozny, {de C. Merschmann},
  and Freitas]{Pereira2011}
Pereira, R.~B., Plastino, A., Zadrozny, B., {de C. Merschmann}, L.~H., and
  Freitas, A.~A.
\newblock Lazy attribute selection: Choosing attributes at classification time.
\newblock \emph{Intelligent Data Analysis}, 15\penalty0 (5):\penalty0 715--732,
  August 2011.

\bibitem[Tacutu et~al.(2013)Tacutu, Craig, Budovsky, Wuttke, Lehmann,
  Taranukha, Costa, Fraifeld, and {de Magalh\~{a}es}]{Tacutu2013}
Tacutu, R., Craig, T., Budovsky, A., Wuttke, D., Lehmann, G., Taranukha, D.,
  Costa, J., Fraifeld, V.~E., and {de Magalh\~{a}es}, J.~P.
\newblock Human ageing genomic resources: Integrated databases and tools for
  the biology and genetics of ageing.
\newblock \emph{Nucleic Acids Research}, 41\penalty0 (D1):\penalty0
  D1027--D1033, January 2013.

\bibitem[{The Gene Ontology Consortium}(2000)]{GO2000}
{The Gene Ontology Consortium}.
\newblock Gene {O}ntology: tool for the unification of biology.
\newblock \emph{Nature Genetics}, 25\penalty0 (1):\penalty0 25--29, May 2000.

\bibitem[Wan(2015)]{WANPHD}
Wan, C.
\newblock \emph{{Novel Hierarchical Feature Selection Methods for
  Classification and Their Application to Datasets of Ageing-Related Genes}}.
\newblock PhD thesis, University of Kent, 2015.

\bibitem[Wan(2016)]{Wan2016AIMatters}
Wan, C.
\newblock {Novel hierarchical feature selection algorithms for predicting
  genes' aging-related function}.
\newblock \emph{AI Matters}, 2:\penalty0 23--24, 2016.

\bibitem[Wan \& Freitas(2013)Wan and Freitas]{Wan2013}
Wan, C. and Freitas, A.~A.
\newblock Prediction of the pro-longevity or anti-longevity effect of
  \emph{Caenorhabditis Elegans} genes based on {B}ayesian classification
  methods.
\newblock In \emph{Proc. of {IEEE} International Conference on Bioinformatics
  and Biomedicine ({BIBM} 2013)}, pp.\  373--380, Shanghai, China, December
  2013.

\bibitem[Wan \& Freitas(2015)Wan and Freitas]{WanACMBCB2015}
Wan, C. and Freitas, A.~A.
\newblock Two methods for constructing a gene ontology-based feature selection
  network for a {B}ayesian network classifier and applications to datasets of
  aging-related genes.
\newblock In \emph{Proc. of the Sixth ACM Conference on Bioinformatics,
  Computational Biology and Health Informatics (ACM-BCB 2015)}, pp.\  27--36,
  Atlanta, USA, Sept. 2015.

\bibitem[Wan et~al.(2015)Wan, Freitas, and {de Magalh\~{a}es}]{Wan2014}
Wan, C., Freitas, A.~A., and {de Magalh\~{a}es}, J.~P.
\newblock Predicting the pro-longevity or anti-longevity effect of model
  organism genes with new hierarchical feature selection methods.
\newblock \emph{IEEE/ACM Transactions on Computational Biology and
  Bioinformatics}, 12\penalty0 (2):\penalty0 262--275, March 2015.

\bibitem[Zhang \& Ling(2001)Zhang and Ling]{Zhang2001TAN}
Zhang, H. and Ling, C.~X.
\newblock An improved learning algorithm for augmented naive bayes.
\newblock \emph{Advances in Knowledge Discovery and Data Mining},
  2035:\penalty0 581--586, April 2001.

\end{thebibliography}
\bibliographystyle{icml2016}

\end{document}